\documentclass[sort,compress,11pt]{elsarticle}

\usepackage{amssymb}
\usepackage{amsthm}
\usepackage{amsmath, esint}
\usepackage{mathtools}
\usepackage{mathrsfs}
\usepackage[ruled, lined, linesnumbered, longend]{algorithm2e}
\usepackage{array}
\usepackage{booktabs}
\usepackage{multirow}
\usepackage{listings}
\usepackage{tabu}
\usepackage{enumerate}
\usepackage{lineno}
\usepackage{fullpage}
\usepackage{xcolor}
\usepackage[colorinlistoftodos]{todonotes}
\usepackage[colorlinks=true]{hyperref}
\usepackage{url}
\usepackage{xurl}
\usepackage{textcomp}

\usepackage{graphicx}
\usepackage{subcaption}

\usepackage{makecell}
\usepackage[export]{adjustbox}
\usepackage{caption}
\usetikzlibrary{shapes}
\biboptions{numbers,comma,round,square}
\usepackage[para,online,flushleft]{threeparttable}

\newcolumntype{L}{!{\color{gray}\vline}} 

\usepackage{orcidlink}

\definecolor{myblue}{RGB}{63, 144, 218}
\definecolor{myorange}{RGB}{221, 132, 82}
\definecolor{mygreen}{RGB}{85, 168, 104}
\definecolor{myyellow}{RGB}{255, 169, 14}
\definecolor{myred}{RGB}{189, 31, 1}

\graphicspath{ {./figs/} }

\journal{Elsevier}
\DontPrintSemicolon
\definecolor{orcidlogocol}{HTML}{A6CE39}
\begin{document}
\makeatletter
\def\ps@pprintTitle{%
  \let\@oddhead\@empty
  \let\@evenhead\@empty
  \let\@oddfoot\@empty
  \let\@evenfoot\@oddfoot
}
\makeatother
\begin{frontmatter}


\title{High-Fidelity Industrial Crash Dynamics Prediction via Geometry-Aware Operator Learning with Memory-Efficient Low-Rank Attention}

\author[ndAME]{Deepak Akhare\orcidlink{0000-0001-7479-3726}}
\address[ndAME]{Department of Aerospace and Mechanical Engineering, University of Notre Dame, Notre Dame, IN}

\author[NV]{Mohammad Amin Nabian}
\address[NV]{NVIDIA, San Jose, CA}

\author[NV]{Corey Adams}

\author[GM]{Sudeep Chavare}
\address[GM]{General Motors, Detroit, MI}

\author[NV]{Sanjay Choudhry}


\begin{abstract}
Automotive crashworthiness optimization remains a safety-critical challenge, requiring the management of large-scale nonlinear structural deformations and energy dissipation through iterative, high-fidelity simulations. While traditional finite element solvers are computationally prohibitive, emerging operator learning frameworks provide rapid surrogate predictions; however, applying them to industrial-scale crash analysis—where complex geometry, contact nonlinearities, and rapidly evolving transient deformation coexist—remains an open challenge. In this paper, we demonstrate that the GeoTransolver \cite{adams2025geotransolver} framework
\footnote{The implementation code is available at: \url{https://github.com/NVIDIA/physicsnemo/blob/main/examples/structural_mechanics/crash/README.md}} provides a viable solution for accurate, high-fidelity crash dynamics prediction at industrial scale. Benchmarked on complex bumper beam and Full-Vehicle crash datasets, GeoTransolver captures multi-scale geometric context and accurately resolves plastic deformation patterns as well as acceleration profiles at critical occupant locations. Beyond the architecture itself, we propose and systematically evaluate a suite of temporal prediction recipes—including one-shot, time-conditional, and autoregressive rollout strategies—demonstrating that the one-shot approach achieves state-of-the-art accuracy with significantly reduced training overhead and inference latency. As a secondary contribution, we introduce a Fast Low-rank Attention Routing Engine (FLARE)-based modification to the GeoTransolver attention backbone that reduces memory overhead by approximately $2\times$ while further improving predictive accuracy for $O(N)$ long-range, high-frequency transients, preserving the geometry-aware cross-attention strengths of the base framework. Our results highlight the practical viability of geometry-aware operator learning for high-fidelity surrogate modeling of complex, safety-critical automotive dynamics.
\end{abstract}


\end{frontmatter}

\section{Introduction}

Modern automotive engineering emphasizes vehicle safety as a core design objective alongside traditional performance and efficiency metrics. In this context, crashworthiness---the ability of a vehicle structure to protect its occupants by governing deformation under impact---is a decisive factor in consumer safety ratings and regulatory compliance\cite{du2004vehicle, hershman2001us}. High crashworthiness is achieved not through maximum stiffness, but through the deliberate engineering of controlled deformation mechanisms that absorb kinetic energy while limiting acceleration levels and cabin intrusion \cite{ozcan2021analysis, li2017crashworthiness, wang2018crashworthiness}. Designing such safety-critical systems requires evaluating the influence of geometry, material distribution, and structural layout on crash response, making crashworthy vehicle development an inherently complex and iterative process.

In practice, this process relies heavily on high-fidelity Finite Element Analysis (FEA) to simulate the complex, nonlinear deformation patterns that arise during impact events across different structural configurations~\cite{sun2011crashworthiness, liu2014simulation, sequeira2020fem, karapetkov2022dynamic, hickey2017finite, bendjaballah2025modeling}. However, the high computational cost of traditional FEA solvers severely limits the number of designs that can be explored, making large-scale multi-objective optimization computationally prohibitive~\cite{fang2005comparative, thel2025accelerating, guo2023machine}. This challenge has motivated increasing interest in AI-driven surrogate modeling, which seeks to accelerate simulation-intensive design workflows while maintaining predictive fidelity. 
Recent studies have demonstrated the effectiveness of data-driven surrogates for a range of nonlinear structural mechanics problems, including load--displacement prediction up to ductile fracture \cite{dey2025data}, stress-field reconstruction in viscoplastic materials \cite{khorrami2023artificial}, large-deformation elastic--plastic response \cite{tao2025ladeep}, displacement prediction from sparse sensing data \cite{hashemi2025physics}, drivetrain dynamics prediction \cite{koutsoupakis2023drivetrain}, and surrogate-assisted material and structural optimization \cite{wang2021surrogate, toda2023surrogate, altoyuri2024plastic, huang2025physics}. More broadly, recent reviews have highlighted the rapid expansion of machine learning for structural and engineering design, simulation acceleration, and performance prediction \cite{ao2025artificial, shao2023accelerating, bakhshan2026ai}. In addition, mesh-free and graph-based learning approaches have shown promise for structural systems with irregular discretizations, geometry-dependent behavior, and plastic deformation processes \cite{hoffer2021mesh, li2025new, st2026graph}.

Despite these advances, the application of AI surrogates to high-fidelity vehicle crash analysis remains comparatively limited, largely because crash events involve a particularly challenging combination of complex geometry, different materials, contact nonlinearities, and rapidly evolving transient deformation. These challenges call for learning frameworks that can capture both localized deformation mechanisms and long-range structural interactions efficiently. In this regard, recent advances in scientific machine learning have introduced operator learning methods that aim to approximate mappings between infinite-dimensional function spaces, enabling rapid surrogate prediction without repeatedly solving the underlying governing equations \cite{lu2021learning, li2020fourier, nabian2025automotive, gao2022multi}. Such approaches provide a promising foundation for crashworthiness modeling, where accurate and efficient prediction of structural response across varying designs and impact conditions is essential \cite{gao2022multi, guo2023machine}.

Among these, transformer-based architectures such as OFormer \cite{li2022oformer} and GNOT \cite{hao2023gnot} have shown promise in handling heterogeneous inputs, though they often struggle to scale to the massive mesh densities required for full-vehicle crash analysis. Recent models such as Transolver \cite{wu2024transolver} and GeoTransolver \cite{adams2025geotransolver} address this challenge through learnable physical slices, achieving linear complexity while capturing global context efficiently. In particular, GeoTransolver couples linear-complexity physics-attention with multi-scale geometric inductive biases, making it a strong candidate for crash dynamics, where rapid propagation of deformation and force through structurally diverse regions demands both global communication and geometric grounding. In this work, we leverage GeoTransolver as a viable, industrial-scale surrogate for high-fidelity crash dynamics prediction and benchmark it on complex bumper beam and Full-Vehicle datasets. In addition, motivated by the memory footprint of physics-attention at industrial mesh densities, we explore a Fast Low-rank Attention Routing Engine (FLARE)-based \cite{puri2025flare} modification to the GeoTransolver attention backbone that retains the geometry-aware cross-attention to the global context but routes self-attention through a low-rank latent bottleneck. This modification reduces memory overhead by approximately $2\times$ while further improving predictive accuracy for the long-range, high-frequency transients characteristic of crash events.

Our key contributions are:
\begin{itemize}
    \item We demonstrate that GeoTransolver \cite{adams2025geotransolver} provides a viable, industrial-scale surrogate for high-fidelity crash dynamics, achieving accurate predictions of position, velocity, and acceleration—including at critical occupant locations—on complex bumper beam and Full-Vehicle crash benchmarks.
    \item We evaluate and compare various temporal dynamics prediction strategies---\textbf{one-shot, time-conditional, and autoregressive rollout}---for highly nonlinear crash events, and identify recipes that achieve state-of-the-art accuracy with significantly reduced training and inference cost.
    \item We propose a Fast Low-rank Attention Routing Engine (FLARE)-based \cite{puri2025flare} modification to the GeoTransolver attention backbone that reduces memory overhead by approximately $2\times$ while further improving predictive accuracy, by retaining the geometry-aware cross-attention to the global context and routing self-attention through a low-rank latent bottleneck.
\end{itemize}

\section{Background and Related Work}
To perform spatial attention, the discretized $N$ points, mesh cells, or nodes are typically treated as a sequence of inputs to the transformer. The scaling of transformer-based operator learning to complex physical domains is primarily constrained by the $O(N^{2})$ complexity of standard self-attention:
\begin{equation}
    \text{Attention}(Q, K, V) = \text{softmax}\left(\frac{QK^T}{\sqrt{d_k}}\right)V,
\end{equation}
where $Q$, $K$, and $V$ represent the query, key, and value matrices obtained by linear projections of the input sequence. This quadratic scaling becomes prohibitive for large-scale meshes or point clouds. To maintain high fidelity while ensuring computational efficiency, recent research has focused on linear-complexity attention variants that either group spatial features into learnable physical tokens or route global communication through latent bottlenecks \cite{wu2024transolver, puri2025flare, adams2025geotransolver}. This section reviews the core architectures that form the basis of our work: Transolver, FLARE, and GeoTransolver.

\subsection{Transolver}
Transolver \cite{wu2024transolver} addresses the challenges of modeling PDEs on general geometries by introducing the \emph{Physics-Attention} mechanism. Rather than applying attention directly over all mesh points, Transolver first decomposes the input feature map $X \in \mathbb{R}^{N \times C}$ into $M$ learnable slices:
\begin{equation}
    S = \frac{\text{softmax}(P(X))}{\text{sum}(\text{softmax}(P(X)), \text{dim}=0)}  \in \mathbb{R}^{N \times M},
\end{equation}
where $P(\cdot)$ projects pointwise features into slice weights, and $S_{ij}$ denotes the degree to which point $i$ belongs to slice $j$. These slices are then aggregated into physics-aware tokens
\begin{equation}
    T =  S^T X \in \mathbb{R}^{M \times C},
\end{equation}
so that each token summarizes a latent physical state shared by a subset of mesh points. Attention is subsequently performed among these encoded tokens:
\begin{equation}
    \hat{T} = \text{Attention}\left(TW_Q, TW_K, TW_V \right),
\end{equation}
and the updated token features are mapped back to the original mesh through deslicing,
\begin{equation}
    Y = S\hat{T}.
\end{equation}
This design gives Transolver linear complexity with respect to the number of mesh points when the number of slices is fixed, while biasing the model toward learning interactions among latent physical states rather than raw mesh nodes \cite{wu2024transolver}.

\subsection{FLARE: Fast Low-rank Attention Routing Engine}
In a related line of linear-complexity research, the Fast Low-rank Attention Routing Engine (FLARE) \cite{puri2025flare} provides a distinct formulation for spatial attention through latent token routing. Instead of learning slices and then applying self-attention among them, FLARE introduces a set of $M \ll N$ learnable latent queries $Q_\ell \in \mathbb{R}^{M \times C}$ and performs attention in two cross-attention stages. Given input $X$,
FLARE first encodes the $N$ input tokens into $M$ latent tokens:
\begin{equation}
    Z = \text{softmax}(Q_\ell (XW_K)^T)XW_V \in \mathbb{R}^{M \times C},
\end{equation}
and then decodes the latent information back to the input resolution:
\begin{equation}
    Y = \text{softmax}((XW_K)Q_\ell^T)Z \in \mathbb{R}^{N \times C}.
\end{equation}
Equivalently, this can be written as $Y = W_{\text{dec}} W_{\text{enc}} V,$ with $ W_{\text{enc}} = \text{softmax}(Q_\ell (XW_K)^T) \in \mathbb{R}^{M \times N},$ $W_{\text{dec}} = \text{softmax}((XW_K)Q_\ell^T) \in \mathbb{R}^{N \times M},$ which yields $W = W_{\text{dec}}W_{\text{enc}} \in \mathbb{R}^{N \times N}, \qquad \text{rank}(W) \le M.$
Thus, unlike Transolver, FLARE does not perform latent self-attention among compressed tokens; instead, it realizes token mixing entirely through an encode--decode factorization, yielding a low-rank global routing mechanism with $O(NM)$ complexity \cite{puri2025flare}.

\subsection{GeoTransolver}
GeoTransolver \cite{adams2025geotransolver} builds upon the Transolver framework by integrating multi-scale geometric inductive biases to improve robustness on irregular domains. Central to this architecture is the Geometry-Aware Latent Embeddings (GALE) attention mechanism, which pairs physics-aware self-attention on learned state slices with cross-attention to a shared geometry and boundary-condition context. A key component is the context projector, which processes geometry, global parameters, and boundary conditions to generate a unified context. This projector first augments the input points with multi-scale geometric features derived from ball queries—an approach inspired by the DoMINO \cite{ranade2025domino} framework to balance local fidelity with global coupling. The projector then encodes these features into a shared latent space, which is injected into every transformer block to mitigate representation drift. This architecture allows GeoTransolver to excel in complex CAE benchmarks such as DrivAerML, Luminary SHIFT-SUV, and Luminary SHIFT-Wing by ensuring persistent grounding to the physical domain.

\section{A Memory-Efficient FLARE-Based Modification to GeoTransolver}
\label{sec:geoflare}
\begin{figure}
    \centering
    \includegraphics[width=\linewidth]{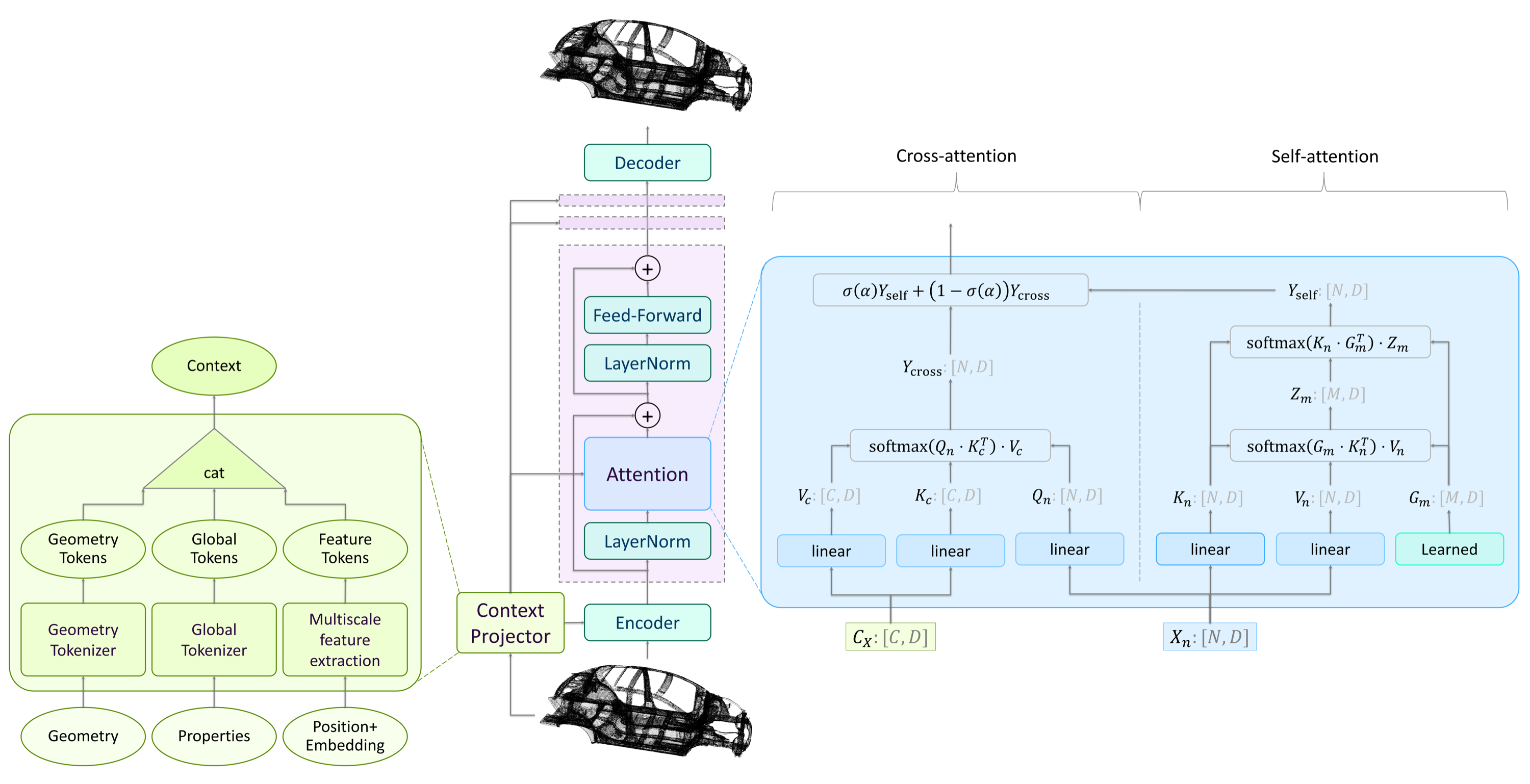}
    \caption{GeoTransolver with FLARE model architecture. Left shows the geometry- and global-state-aware context projector used to augment input points and encode a sample context, and Right shows the attention block that performs the FLARE-based self-attention alongside cross-attention over the context.
    }
\label{fig:GeoFlare}
\end{figure}

As a secondary contribution, we propose a memory-efficient modification to the GeoTransolver attention backbone that replaces its physics-attention self-attention with a FLARE-based low-rank routing mechanism. We refer to this modified variant as \emph{GeoTransolver with FLARE} (\emph{GeoTS-FLARE} in tables). This modification reduces memory overhead by approximately $2\times$ relative to the baseline GeoTransolver, while further improving predictive accuracy by routing global communication through a more expressive low-rank latent bottleneck rather than through spatial slicing. Crucially, unlike the original FLARE formulation, our design retains the multi-scale global context projector from the base GeoTransolver to augment input points and provides a robust geometric foundation via subsequent cross-attention to this global context. The resulting architecture—combining a geometry- and global-state-aware context projector with hybrid self- and cross-attention—is illustrated in Fig.~\ref{fig:GeoFlare}.

\subsection{Self-attention}
The self-attention mechanism in GeoTransolver with FLARE is powered by a FLARE-based engine, which achieves better-predicting linear complexity by routing information through a low-rank latent bottleneck. Specifically, this inspired mechanism introduces a set of $M$ learnable latent tokens (or global queries), where $M \ll N$, to mediate global communication across the input sequence. The operation is divided into encoding and decoding stages. In the encoding stage, learnable global queries $G_m \in \mathbb{R}^{M \times D}$ aggregate information from the input point features $X_n \in \mathbb{R}^{N \times D}$. The key and value representations for this stage are derived as linear projections of the input: $K_n = l(X_n)$ and $V_n = l(X_n)$. This process projects the $N$ input points into a compact latent space represented by $M$ embedding vectors $Z_m \in \mathbb{R}^{M \times D}$, as defined in Eq.~\ref{self_att_enc}. In the decoding stage, these latent representations are projected back to the original point space (Eq.~\ref{self_att_dec}), effectively performing global token mixing with $O(NM)$ complexity.

\begin{subequations}
\begin{equation}
    Z_m = \text{softmax}(G_m\cdot K_n^T)\cdot V_n
    \label{self_att_enc}
\end{equation}    
\begin{equation}
    Y_\text{self} = \text{softmax}(K_n \cdot G_m^T)\cdot Z_m
    \label{self_att_dec}
\end{equation}
\end{subequations}

\subsection{Cross-attention}
The cross-attention mechanism provides essential geometric grounding by allowing the point cloud features to query a multi-scale global context. This context $C_X \in \mathbb{R}^{C \times D}$ is generated by the context projector, which integrates geometry, global parameters, and boundary conditions into a unified representation. The projector first augments the input points with multi-scale geometric features derived from local neighborhood ball queries. These features are then encoded into a shared latent space that remains persistent across all transformer blocks to prevent representation drift. For the cross-attention operation, the point cloud features $X_n$ serve as queries $Q_n = l(X_n)$, while the context $C_X$ provides the keys $K_c = l(C_X)$ and values $V_c = l(C_X)$:
\begin{equation}
    Y_\text{cross} = \text{softmax}(Q_n \cdot K_c^T)\cdot V_c
\end{equation}

The final output of the attention block is an adaptive mixture of the self-attention and cross-attention outputs:
\begin{equation}
    Y_\text{attention} = (1-\sigma(\alpha))Y_\text{cross} + \sigma(\alpha) Y_\text{self}
\end{equation}
where $\alpha$ is a learnable parameter initialized at zero that allows the model to dynamically balance global communication with geometric constraints.

\section{Temporal Dynamics Prediction Strategies}
\label{sec:time_int}
The simulation data in this study are uniformly discretized in time, where the solution states $\mathbf{x}^{(t)} \in \mathbb{R}^{N\times D}$ are defined at discrete intervals $t \in \{ i\Delta t \}_{i=0}^T$. Here, $\Delta t$ denotes the constant temporal step across a total horizon of $T$ increments, and $\mathbf{f} \in \mathbb{R}^{N\times F}$ represents auxiliary design features provided to the model. To capture the complex evolution of a crash event, we evaluate four primary formulations for predicting the estimated states $\hat{\mathbf{x}}^{(t)}$.

\subsection{Autoregressive (AR) Rollout} 
 
The AR strategy predicts the system state at time $t$ based on its preceding history:
\begin{equation}
    \hat{\mathbf{x}}^{(n\Delta t)} = f(\{\hat{\mathbf{x}}^{(i\Delta t)} \}_{i=0}^{n-1}, \textbf{f}).
\end{equation}

Specifically, inspired by ordinary differential equation (ODE) systems, we utilize a neural network to predict the instantaneous acceleration $\hat{\mathbf{a}}^{(t)}$ based on the two previous states. These predictions are then integrated using an explicit Euler scheme to derive the velocity $\hat{\mathbf{v}}^{(t)}$ and the subsequent deformed position $\hat{\mathbf{x}}^{(t)}$:
\begin{subequations}
\begin{equation}
    \hat{\mathbf{a}}^{(t)} =  NN(\hat{\mathbf{x}}^{(t-\Delta t)}, \hat{\mathbf{x}}^{(t-2\Delta t)}, \textbf{f}),
\end{equation}
\begin{equation}
    \hat{\mathbf{v}}^{(t)} = \hat{\mathbf{v}}^{(t-\Delta t)} + \Delta t \cdot \hat{\mathbf{a}}^{(t)},
\end{equation}
\begin{equation}
    \hat{\mathbf{x}}^{(t)} = \hat{\mathbf{x}}^{(t-\Delta t)} + \Delta t \cdot \hat{\mathbf{v}}^{(t)}.
\end{equation}
\end{subequations}

During training, the entire trajectory is unrolled from the initial condition $\mathbf{x}^{(0)}$ such that $\mathbf{x}^{(0)} \rightarrow \hat{\mathbf{x}}^{(\Delta t)} \rightarrow \dots \rightarrow \hat{\mathbf{x}}^{(T)}$. The loss is computed over the full sequence using the $L_2$ norm:
\begin{equation}
    L = \sum_{i=1}^{T} ||\hat{\mathbf{x}}^{(i\Delta t)} - {\mathbf{x}}^{(i\Delta t)}||_2.
\end{equation}

\subsection{Teacher Forcing}
Teacher forcing is a training variant of the AR rollout where the model is "corrected" at each step by using the ground truth states $\{{\mathbf{x}}^{(i\Delta t)} \}_{i=0}^{n-1}$ as input rather than its own previous prediction:
\begin{equation}
    \hat{\mathbf{x}}^{(n\Delta t)} = f(\{{\mathbf{x}}^{(i\Delta t)} \}_{i=0}^{n-1}, \textbf{f}).
\end{equation}
This strategy prevents error accumulation and divergence during the training phase, ensuring more stable gradient updates. While it follows the same integration logic as the AR rollout, the reliance on ground truth states decouples the model's current prediction from past inaccuracies.

\subsection{One-shot Prediction}
In the One-shot formulation, the model maps the initial condition $\mathbf{x}^{(0)}$ directly to the complete temporal sequence in a single forward pass:
\begin{subequations}
\begin{equation}
    \hat{\mathbf{y}} = f(\mathbf{x}^{(0)}, \textbf{f}),
\end{equation}
\begin{equation}
    [\hat{\mathbf{x}}^{(0)}, \cdots, \hat{\mathbf{x}}^{(i\Delta t)}, \cdots, \hat{\mathbf{x}}^{(T\Delta t)}] = \hat{\mathbf{y}},
\end{equation}
\end{subequations}
Here, the model output $\hat{\mathbf{y}} \in \mathbb{R}^{N \times (D \cdot T)}$ treats all timesteps as a concatenated feature vector. This approach enables fully parallelized prediction and bypasses the recursive instability often associated with long-range autoregressive rollouts.

\subsection{Time-conditional Mapping}
Time-conditional mapping treats time as a continuous query coordinate, predicting the state at any target time $t$ as a function of the initial condition and a temporal offset:
\begin{equation}
    \hat{\mathbf{x}}^{t} = f(\mathbf{x}^{(0)}, \textbf{f}, t).
\end{equation}
By reformulating the problem as a continuous mapping, the model can query states at any point within the simulation horizon without iterative stepping.

\begin{figure}
    \centering
    \includegraphics[width=\linewidth]{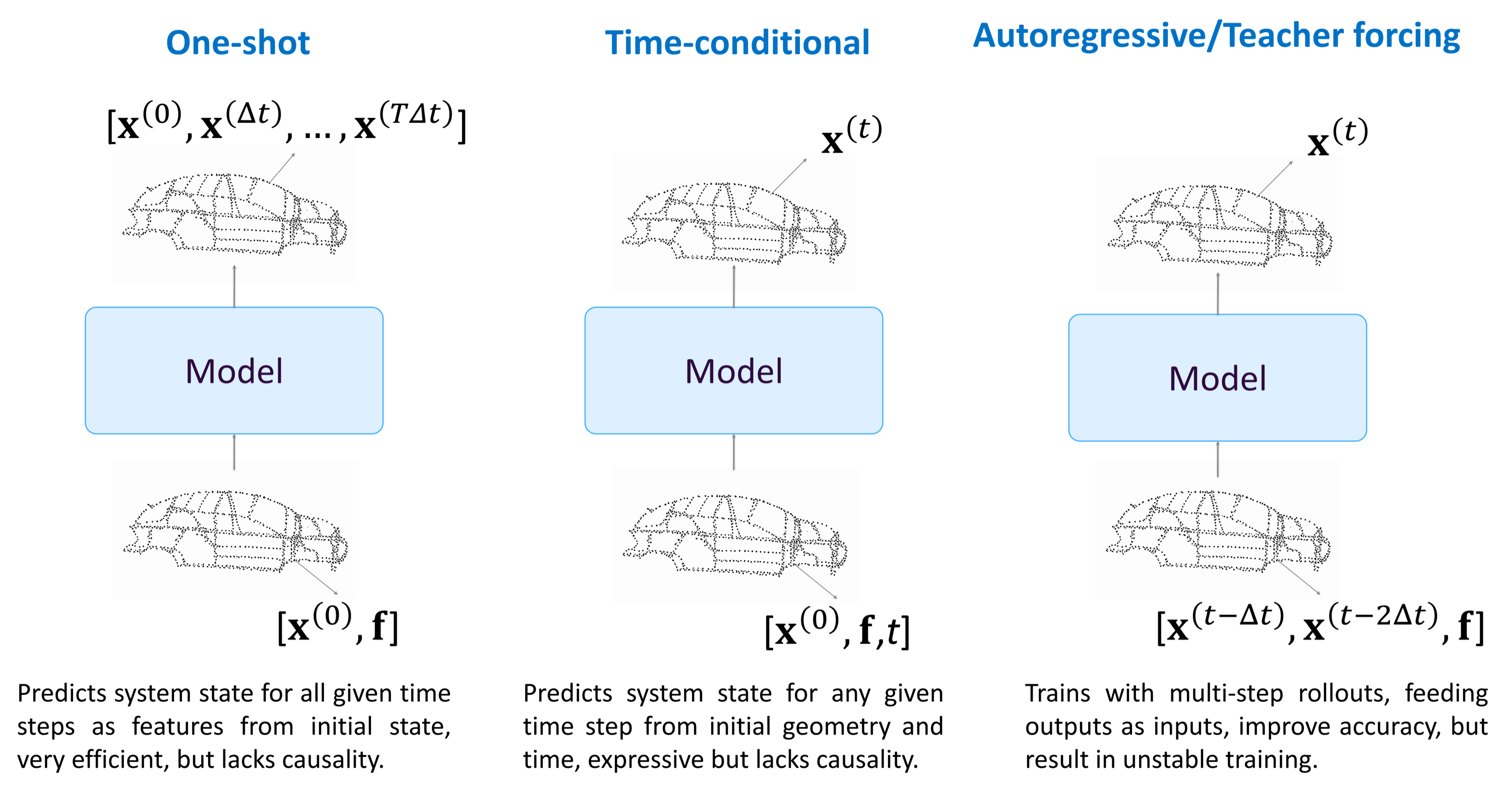}
    \caption{Schematic illustration of temporal dynamics prediction strategies.}
    \label{fig:time_dyn}
\end{figure}

\subsection{Comparison of Temporal Formulations}
While the rollout strategies (AR and teacher forcing) strictly obey physical causality, they are susceptible to cumulative error during inference as inaccuracies in the previous state propagate forward. Conversely, One-shot and time-conditional strategies ignore strict causality but are immune to such drift. Furthermore, the inference cost of the one-shot and time-conditional models is $\mathcal{O}(1)$, whereas the iterative nature of AR and teacher forcing results in an $\mathcal{O}(T)$ computational overhead.

\begin{table}[ht]
\centering
\caption{Comparison of Temporal Dynamics Prediction Strategies.}
\label{tab:strategy_comparison}
\begin{tabular}{@{}lcccc@{}}
\toprule
\textbf{Strategy} & \textbf{Causality} & \textbf{Inference Cost} & \textbf{Error Propagation} & \textbf{Training Stability} \\ \midrule
AR rollout        & Yes                & $\mathcal{O}(T)$        & High                       & Low                    \\
Teacher forcing   & Yes                & $\mathcal{O}(T)$        & High (at Inference)        & High                        \\
One-shot          & No                 & $\mathcal{O}(1)$        & None                       & High                        \\
Time-conditional  & No                 & $\mathcal{O}(1)$        & None                       & High                        \\ \bottomrule
\end{tabular}
\end{table}


\section{Dataset}

\subsection{Bumper Beam Dataset}
The bumper beam dataset is generated from a parametric finite element representation of a bumper beam impact scenario, adapted from the publicly available Altair front-impact bumper reference case and the corresponding OpenRadioss model distribution. Crash simulations are carried out using the explicit finite element solver OpenRadioss, which is suited for nonlinear transient dynamics involving large deformations, contact interactions, and elastoplastic material response. In all simulations, a shell-meshed bumper beam is struck by a cylindrical rigid-wall impactor along the longitudinal axis, and time-resolved nodal displacements, velocities, and contact forces are recorded throughout the event. The model retains the key structural features of the bumper assembly relevant to crash response while remaining computationally tractable, enabling a realistic representation of deformation and energy-absorption modes during impact. To construct a diverse dataset, a Design of Experiments was conducted by jointly varying five input factors: anisotropic geometric scaling of the beam along the X, Y, and Z axes (5 configurations); initial impactor velocity along the longitudinal direction (3 levels: -3, -5, and -7 mm/ms); uniform shell thickness scaling (3 levels: 0.7×, 1.0×, and 1.3× nominal); rigid-wall impactor diameter (fixed at 254 mm); and transverse impactor origin offset along the Y axis (3 positions at 0, 120, and 240 mm). The full-factorial combination yields 135 distinct design configurations. For each configuration, a full crash simulation is executed, producing spatiotemporal fields of structural deformation that characterize the mapping from design parameters to crash performance. The dataset is divided into training (80\%), validation (10\%), and testing (10\%) subsets.

\subsection{Car crash: The Body-in-White (BIW) Dataset}
The dataset used in this work is generated from a high-fidelity finite element representation of a Body-in-White (BIW) vehicle structure, adapted from a publicly available model released by the National Highway Traffic Safety Administration (NHTSA) \cite{NHTSA_CrashModels}. The model consists of approximately 400,000 nodes and 380,000 elements, retaining the key structural components relevant to crash response while maintaining computational tractability.

Crash simulations were carried out using the explicit finite element solver LS-DYNA, which is widely used for nonlinear transient dynamics involving large deformations, complex material behavior, and contact interactions. In all simulations, the vehicle structure undergoes a frontal impact with a rigid barrier at a speed of 56 km/h, consistent with standardised crash-test conditions. The BIW model includes critical structural features such as crush zones, reinforcements, and energy-absorbing members, enabling a realistic representation of deformation modes during impact. To balance resolution and computational cost, simulations are performed on a reduced vehicle subsystem using a mesh size of approximately 10 mm, which is sufficient to capture localized deformation phenomena. Each simulation spans 120 milliseconds of physical time and requires over an hour of wall-clock time on a high-performance computing cluster.

To construct a diverse dataset, a Design of Experiments was conducted by perturbing the thickness of 33 front-end structural components by $\pm20\%$ around their nominal values. This process results in 150 distinct design configurations. For each configuration, a full crash simulation is executed, producing detailed spatiotemporal fields consisting of 25 discrete time steps. These fields include structural deformation and acceleration responses, providing a rich dataset for learning the relationship between design parameters and crash performance. The dataset is divided into training (90\%), validation (5\%), and testing (5\%) subsets.

\begin{figure}
\centering
\begin{minipage}{.5\textwidth}
  \centering
  \includegraphics[width=0.95\linewidth]{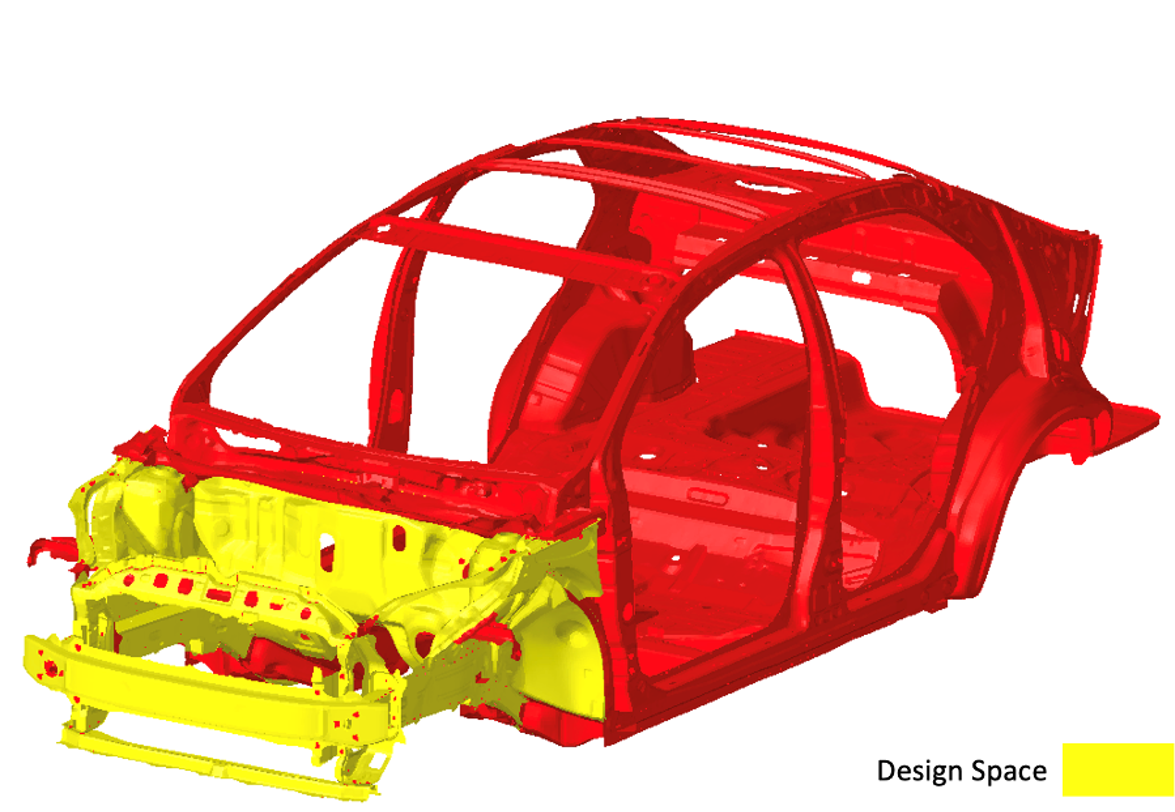}
  \captionof{figure}{Parts included in the design space. }
  \label{fig:Design_space}
\end{minipage}%
\begin{minipage}{.5\textwidth}
  \centering
  \includegraphics[width=0.95\linewidth]{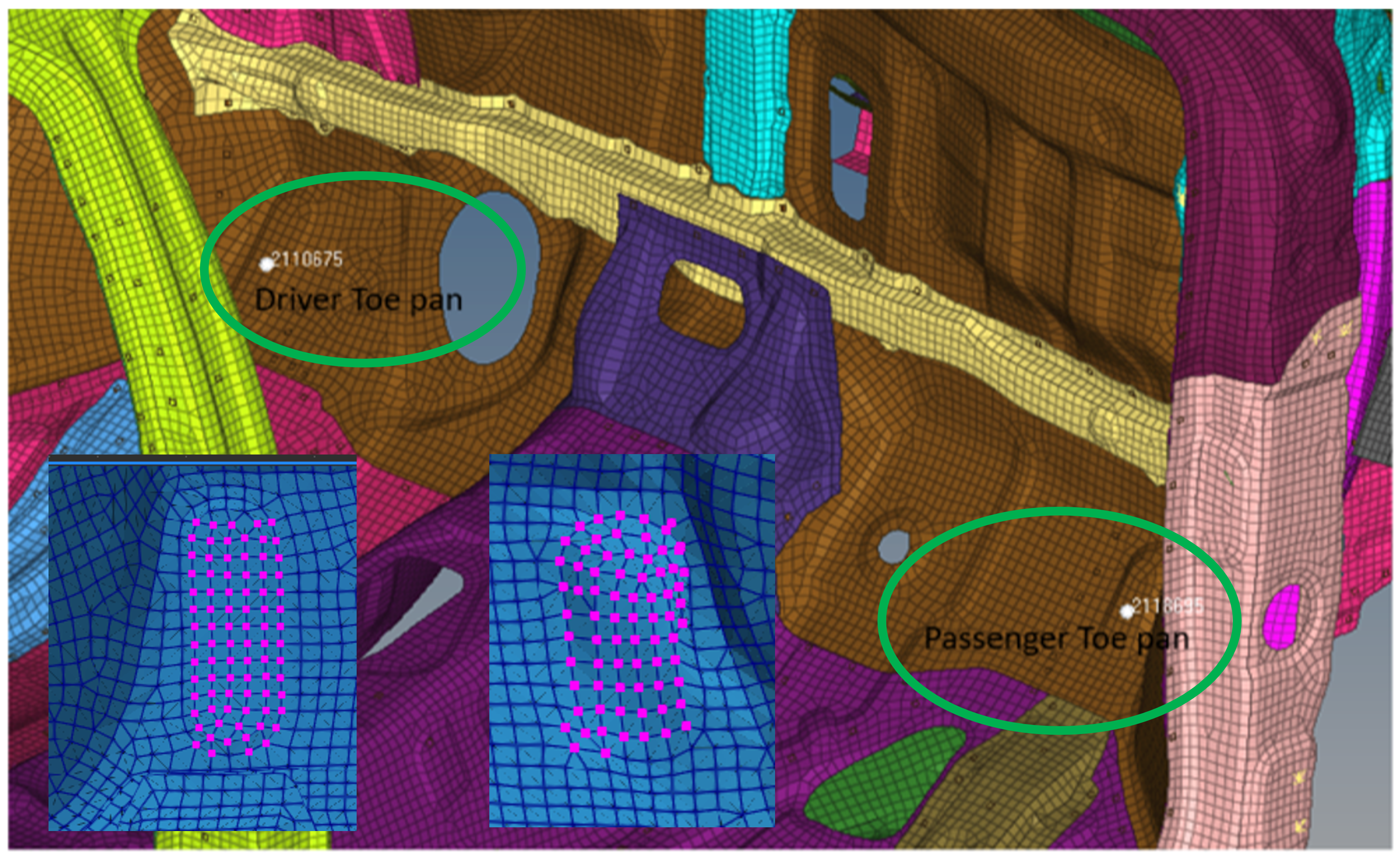}
  \captionof{figure}{Probe points to measure the acceleration at the driver and passenger Toe Pans. }
  \label{fig:Probe_points}
\end{minipage}
\end{figure}

In addition to full-field outputs, localized measurements are extracted at selected probe locations situated near the driver and passenger toe pan regions, as shown in Fig.~\ref{fig:Probe_points}. These locations are particularly sensitive to front-end structural deformation and serve as reliable indicators of occupant safety during frontal impacts. The recorded acceleration histories at these probe points are therefore used as key response quantities for evaluating crashworthiness performance.

\section{Results}

\subsection{Evaluation of Temporal Dynamics Prediction Strategies}
To identify the most effective temporal formulation for crash modeling, we compare the performance of the four strategies defined in Section~\ref{sec:time_int}: autoregressive rollout, teacher forcing, one-shot prediction, and time-conditional mapping. For this comparison, we employ the GeoTransolver as the baseline architecture.
The model configurations are tailored to each dataset: for the bumper beam, we use 128 tokens, 6 layers, 8 heads, a 256-channel dimension, and a neighbor radius of $[0.05, 0.25]$ with $[8, 32]$ neighbors for ball queries; for the car crash dataset, the architecture consists of 128 tokens, 5 layers, 8 heads, and a neighbor radius of $[0.05, 0.25]$ with $[8, 32]$ neighbors for ball queries.
The final validation metrics and the average computational cost (time per epoch) for each recipe on both the bumper beam ($T=50$) and car crash ($T=14$) datasets is summarized in Table~\ref{tab:time_int}.


\begin{table}[!ht]
    \centering
    \begin{tabular}{|c|c c ||c| c c|}
        \hline
        Bumper (T=50) & $t$/epoch & Validation MSE & Car (T=14) & $t$/epoch & Validation MSE \\
        \hline
        One-shot & 1 s & $5.42\times10^{-3}$ & One-shot & 6.4 s & $2.32\times10^{-4}$ \\
        Time-condition & 29 s & $4.12\times10^{-3}$ & Time-condition & 40.8 s & $2.54\times10^{-4}$ \\
        AR-rollout & 37 s & unstable & AR-rollout & 61.6 s & $2.27\times10^{-4}$ \\
        Teacher forcing & 29 s & 0.3 &  &  &  \\
        \hline
    \end{tabular}
    \caption{Quantitative comparison of temporal prediction strategies. We report the final validation MSE and computational efficiency (seconds per epoch) for the bumper beam($T=50$) and car crash ($T=14$) datasets.}
    \label{tab:time_int}
\end{table}

The autoregressive (AR) rollout was found to exhibit significant instability for the bumper beam case ($T=50$). While accuracy is maintained for shorter trajectories in the car crash scenario ($T=14$), the model becomes divergent for extended rollouts ($T=25$), indicating that AR unrolling is inherently unstable due to error accumulation for long-horizon dynamics prediction in crash events. Conversely, the time-conditional formulation achieves the highest predictive accuracy for the bumper beam dataset, whereas its performance on the Full-Vehicle benchmark is comparable to the other strategies with slightly broader error margins. One-shot prediction provides a robust balance, delivering competitive accuracy across both datasets with minimal training overhead.

\subsection{Transformer Architecture Comparative Analysis}
To assess how the choice of attention backbone influences accuracy and memory footprint for industrial-scale crash modeling, we conduct a comparative study between Transolver, the baseline GeoTransolver with physics attention, and the proposed FLARE-based modification to GeoTransolver (denoted GeoTransolver with FLARE).
Following the consistent backbone and multi-scale configurations detailed in the previous section, we utilize a token size of 128 for physics attention in Transolver and GeoTransolver, and 128 global queries for the low-rank attention in GeoTransolver with FLARE.
Each architecture is evaluated across both the bumper beam dataset ($T=50$) and the car crash dataset ($T=25$) using two distinct optimization strategies: the standard Adam optimizer and the Muon optimizer \cite{liu2025muon}, which has demonstrated superior convergence properties in high-dimensional physical modeling.

\begin{table}[!ht]
    \centering
    \begin{tabular}{c|c c L c c}
        \hline
        Test Relative $L^2$ & Bumper(Adam) & Bumper(Muon) & Car(Adam) & Car(Muon) \\
        \hline
        TS & $9.37\times10^{-3}$ & $9.12\times10^{-3}$ & $1.60\times10^{-2}$ & $1.60\times10^{-2}$ \\
        GeoTS & $9.40\times10^{-3}$ & $7.32\times10^{-3}$ & $1.40\times10^{-2}$ & $1.33\times10^{-2}$ \\
        \textbf{GeoTS-FLARE} & $\boldsymbol{8.37\times10^{-3}}$ & $\boldsymbol{6.80\times10^{-3}}$ & $\boldsymbol{1.16\times10^{-2}}$ & $\boldsymbol{8.95\times10^{-3}}$ \\
        \hline
    \end{tabular}
    \caption{Comparative analysis of relative $L^2$ test error for bumper beam and car crash datasets across different model architectures and optimization strategies. TS: Transolver, GeoTS: GeoTransolver, and GeoTS-FLARE: GeoTransolver with FLARE}
    \label{tab:L2_arc_adam_muon}
\end{table}

\begin{table}[!ht]
    \centering
    \begin{tabular}{c|c c c c}
        \hline
         & Training speed($t$/epoch) & No. of parameters \\
        \hline
        TS & $4.8$ s & $3,700,783$ \\
        GeoTS & $7.5$ s & $5,686,504$ \\
        GeoTS-FLARE & $7.1$ s & $5,150,624$ \\
        \hline
    \end{tabular}
    \caption{Comparative analysis of training speed and number of parameters for car crash across different model architectures. TS: Transolver, GeoTS: GeoTransolver, and GeoTS-FLARE: GeoTransolver with FLARE}
    \label{tab:speed_para}
\end{table}

\begin{figure}[!ht]
    \centering
    \includegraphics[width=\linewidth]{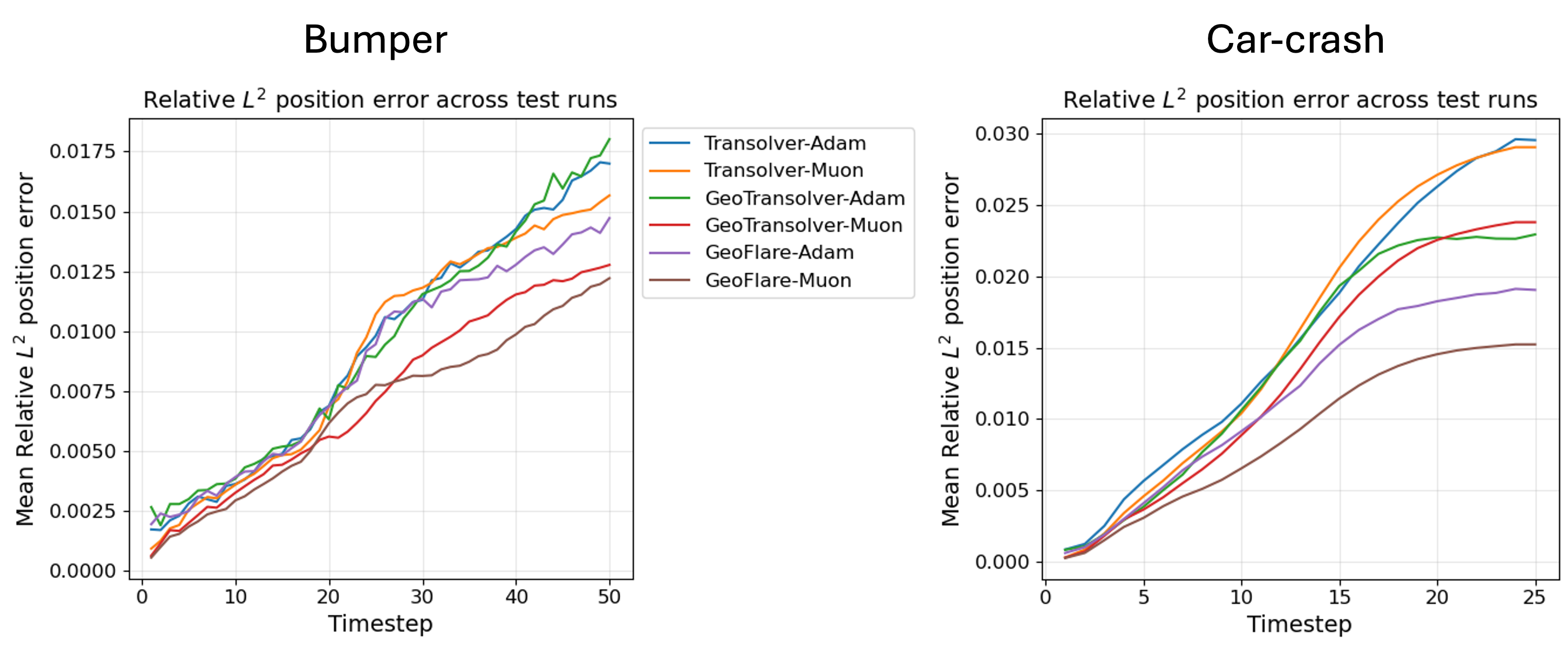}
    \caption{Temporal evolution of the relative $L^2$ test error across the prediction horizon for various model architectures and optimizers.}
    \label{fig:Test_MSE_models_wt}
\end{figure}

Table \ref{tab:L2_arc_adam_muon} summarizes the relative $L^2$ test error for various configurations. The results reveal a consistent performance hierarchy across all benchmarks: Transolver, while efficient, exhibits the highest error rates; GeoTransolver provides significant improvements by incorporating multi-scale geometric inductive biases; and GeoTransolver with FLARE consistently achieves the lowest error, particularly when paired with the Muon optimizer. Specifically, on the car crash dataset, GeoTransolver with FLARE (Muon) reduces the relative $L^2$ error by approximately $33\%$ compared to GeoTransolver and $44\%$ compared to the baseline Transolver. This underscores the superior predictive effectiveness of GeoTransolver with FLARE, demonstrating how substituting the original self-attention with a better-predicting, low-rank latent communication mechanism based on FLARE yields highly accurate approximations while maintaining persistent geometric grounding through cross-attention.
Furthermore, Table~\ref{tab:speed_para} highlights the computational efficiency of these architectures for the car crash surrogate model. While GeoTransolver introduces significant geometric complexity—resulting in a $53\%$ increase in parameters and a $56\%$ slower training cadence compared to the baseline Transolver—the proposed FLARE-based modification reduces this overhead. By replacing physics-attention with a low-rank latent routing mechanism, GeoTransolver with FLARE achieves its superior accuracy with approximately $500{,}000$ fewer parameters and a faster training speed than baseline GeoTransolver. The same modification also reduces peak attention-block memory consumption by approximately $2\times$ relative to baseline GeoTransolver, since the $O(NM)$ encode--decode factorization avoids materializing the full $N\times N$ attention map associated with physics-attention. This memory headroom is particularly valuable for industrial-scale crash simulations, where full-vehicle meshes can otherwise saturate accelerator memory.

The temporal evolution of these test error metrics is visualized in Fig.~\ref{fig:Test_MSE_models_wt}, plotting the mean $L^2$ error across the prediction horizon. GeoTransolver with FLARE maintains superior stability and accuracy throughout the crash event, whereas competing architectures show more pronounced error accumulation over longer intervals.

\begin{table}[!ht]
    \centering
    \begin{tabular}{c c|c c c}
        \hline
         & & Position & Velocity & Acceleration \\
        \hline
        Driver & TS & $2.21\times10^{-3}$ & $8.21\times10^{-1}$ & $5.60\times10^{3}$ \\
        & GeoTS & $1.51\times10^{-3}$ & $5.74\times10^{-1}$ & $3.99\times10^{3}$ \\
        & \textbf{GeoTS-FLARE} & $\boldsymbol{1.01\times10^{-3}}$ & $\boldsymbol{4.38\times10^{-1}}$ & $\boldsymbol{2.99\times10^{3}}$ \\
        \hline
        Passenger & TS & $2.43\times10^{-3}$ & $9.31\times10^{-1}$ & $6.81\times10^{3}$ \\
        & GeoTS & $1.91\times10^{-3}$ & $7.03\times10^{-1}$ & $5.52\times10^{3}$ \\
        & \textbf{GeoTS-FLARE} & $\boldsymbol{1.19\times10^{-3}}$ & $\boldsymbol{5.16\times10^{-1}}$ & $\boldsymbol{3.92\times10^{3}}$ \\
        \hline
    \end{tabular}
    \caption{Driver and passenger MSE for position, velocity, and acceleration on the Full-Vehicle benchmark, comparing GeoTransolver with the FLARE attention modification (GeoTS-FLARE) against the baseline GeoTransolver (GeoTS) and Transolver (TS), all trained with the Adam optimizer.}
    \label{tab:L2_arc_adam_dr_ps}
\end{table}

\begin{table}[!ht]
    \centering
    \begin{tabular}{c c|c c c}
        \hline
         & & Position & Velocity & Acceleration \\
        \hline
        Driver & TS & $2.63\times10^{-3}$ & $8.41\times10^{-1}$ & $2.14\times10^{3}$ \\
        & GeoTS & $1.84\times10^{-3}$ & $6.09\times10^{-1}$ & $1.71\times10^{3}$ \\
        & \textbf{GeoTS-FLARE} & $\boldsymbol{7.18\times10^{-4}}$ & $\boldsymbol{2.71\times10^{-1}}$ & $\boldsymbol{1.27\times10^{3}}$ \\
        \hline
        Passenger & TS & $2.21\times10^{-3}$ & $7.25\times10^{-1}$ & $2.24\times10^{3}$ \\
        & GeoTS & $1.72\times10^{-3}$ & $5.53\times10^{-1}$ & $1.80\times10^{3}$ \\
        & \textbf{GeoTS-FLARE} & $\boldsymbol{6.52\times10^{-4}}$ & $\boldsymbol{2.53\times10^{-1}}$ & $\boldsymbol{1.45\times10^{3}}$ \\
        \hline
    \end{tabular}
    \caption{Driver and passenger MSE for position, velocity, and acceleration on the Full-Vehicle benchmark, comparing GeoTransolver with the FLARE attention modification (GeoTS-FLARE) against the baseline GeoTransolver (GeoTS) and Transolver (TS), all trained with the Muon optimizer.}
    \label{tab:L2_arc_muon_dr_ps}
\end{table}

Furthermore, we evaluate performance on safety-critical metrics for the Full-Vehicle benchmark: the acceleration and velocity profiles at key occupant locations (driver and passenger). Tables \ref{tab:L2_arc_adam_dr_ps} and \ref{tab:L2_arc_muon_dr_ps} detail the Mean Squared Error (MSE) for position, velocity, and acceleration. Across all quantities of interest, GeoTransolver with FLARE demonstrates substantial gains in fidelity. Notably, acceleration predictions---which are highly sensitive to nonlinearities and contact-induced oscillations---show the most significant improvement, further validating the capacity of GeoTransolver with FLARE to resolve the intricate dynamics of automotive impact events.

\subsection{Qualitative Analysis}
In this section, we present a qualitative evaluation of the predictive performance of GeoTransolver across the Full-Vehicle and bumper beam benchmarks. Throughout this section we use the best-performing configuration identified in Section~\ref{sec:geoflare}, namely GeoTransolver with the FLARE attention modification. We focus on the framework's ability to capture the complex, large-scale structural deformations characteristic of automotive impact events, as well as its accuracy in resolving high-fidelity kinematics at critical interior locations.

\begin{figure}[!ht]
    \centering
    \begin{subfigure}{\linewidth}
        \centering
        \includegraphics[width=\linewidth]{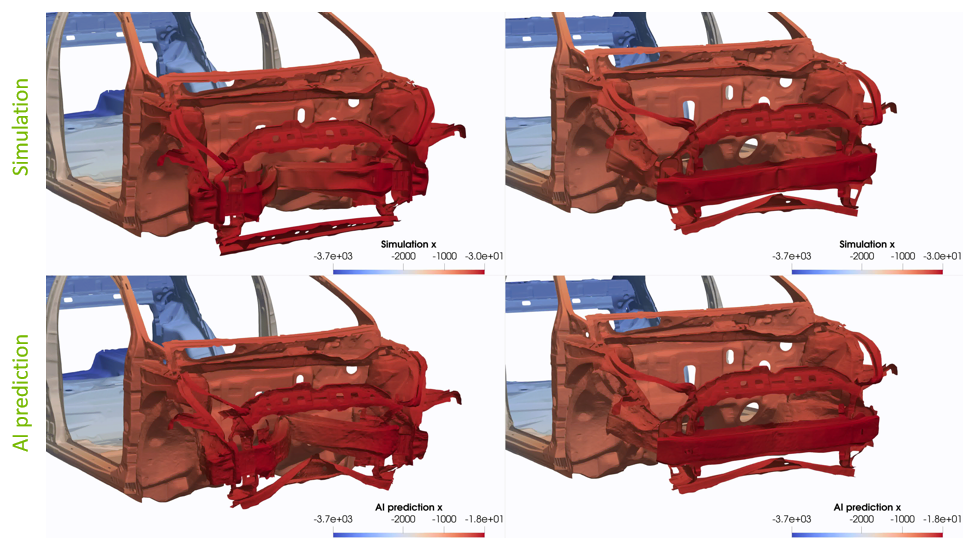}
        \caption{Front view of car crash test samples.}
        \label{fig:car_final_a}
    \end{subfigure}
    \\[2ex] 
    \begin{subfigure}{\linewidth}
        \centering
        \includegraphics[width=\linewidth]{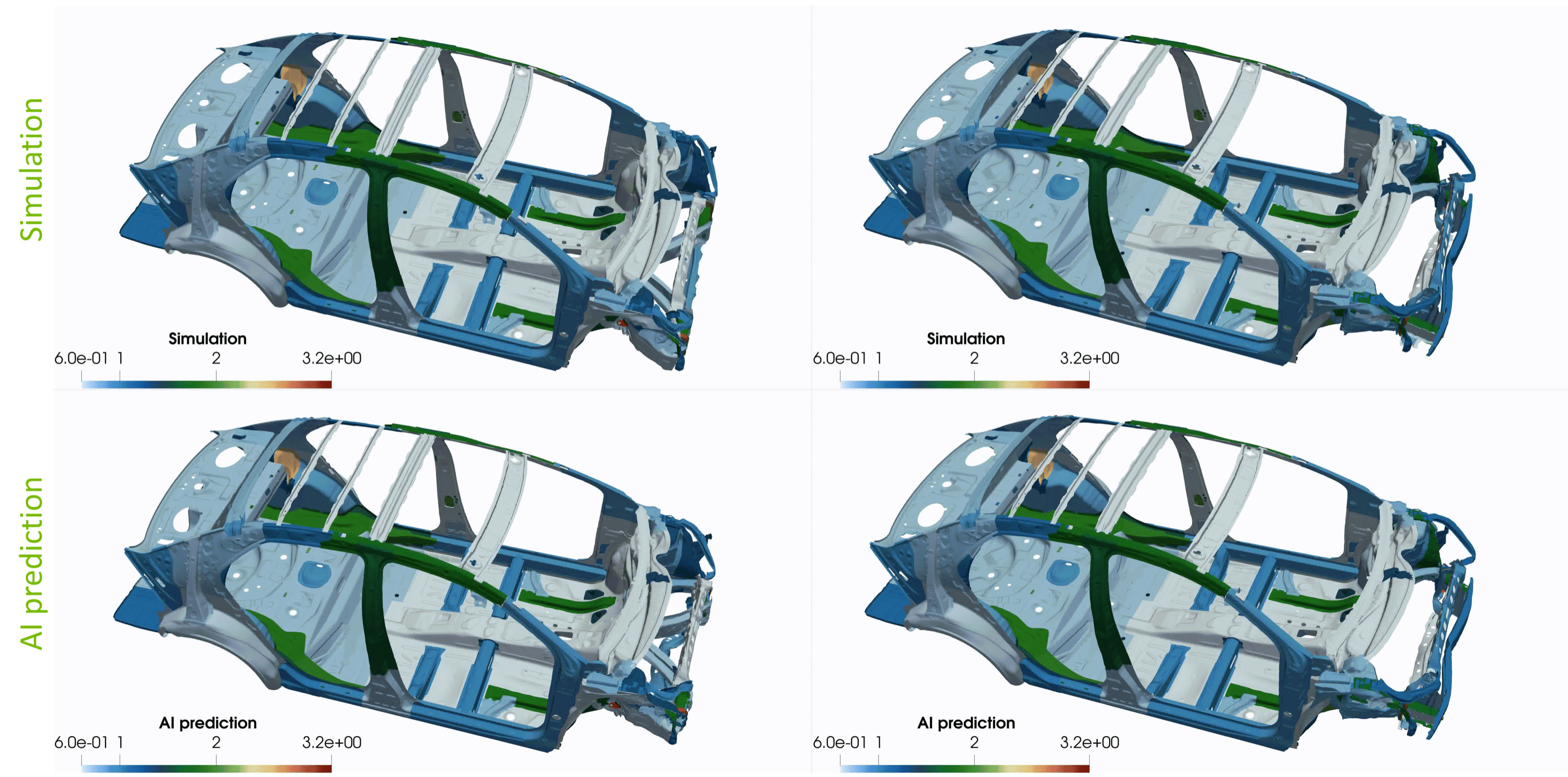}
        \caption{Top view of car crash test samples.}
        \label{fig:car_final_b}
    \end{subfigure}
    \caption{Visual comparison of final deformation for car crash scenarios. In both (a) and (b), the top row shows the ground truth FEA simulation, while the bottom row displays the GeoTransolver prediction.}
    \label{fig:car_final}
\end{figure}

\begin{figure}[!ht]
    \centering
    \begin{subfigure}{\linewidth}
        \centering
        \includegraphics[width=\linewidth]{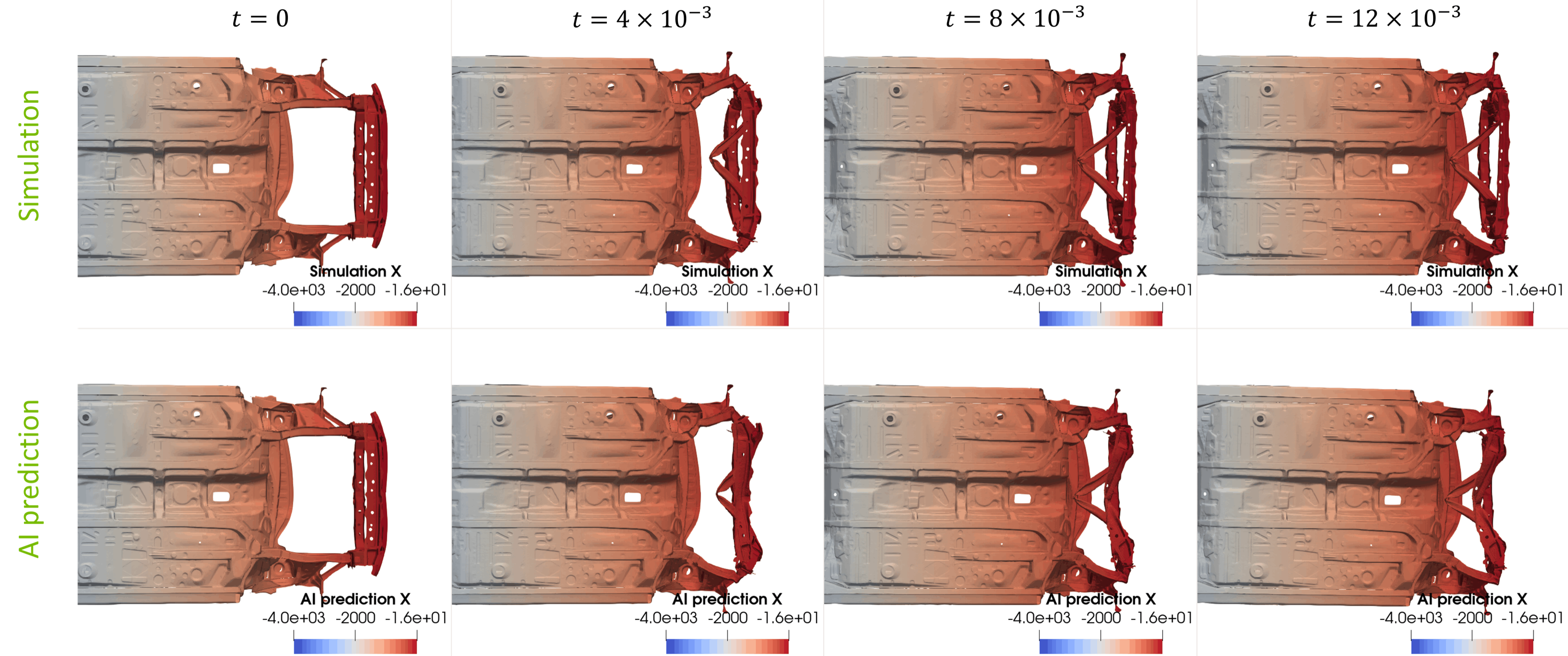}
        \caption{Car crash on test geometry deformation with time.}
        \label{fig:car_wtime_a}
    \end{subfigure}
    \\[2ex] 
    \begin{subfigure}{\linewidth}
        \centering
        \includegraphics[width=\linewidth]{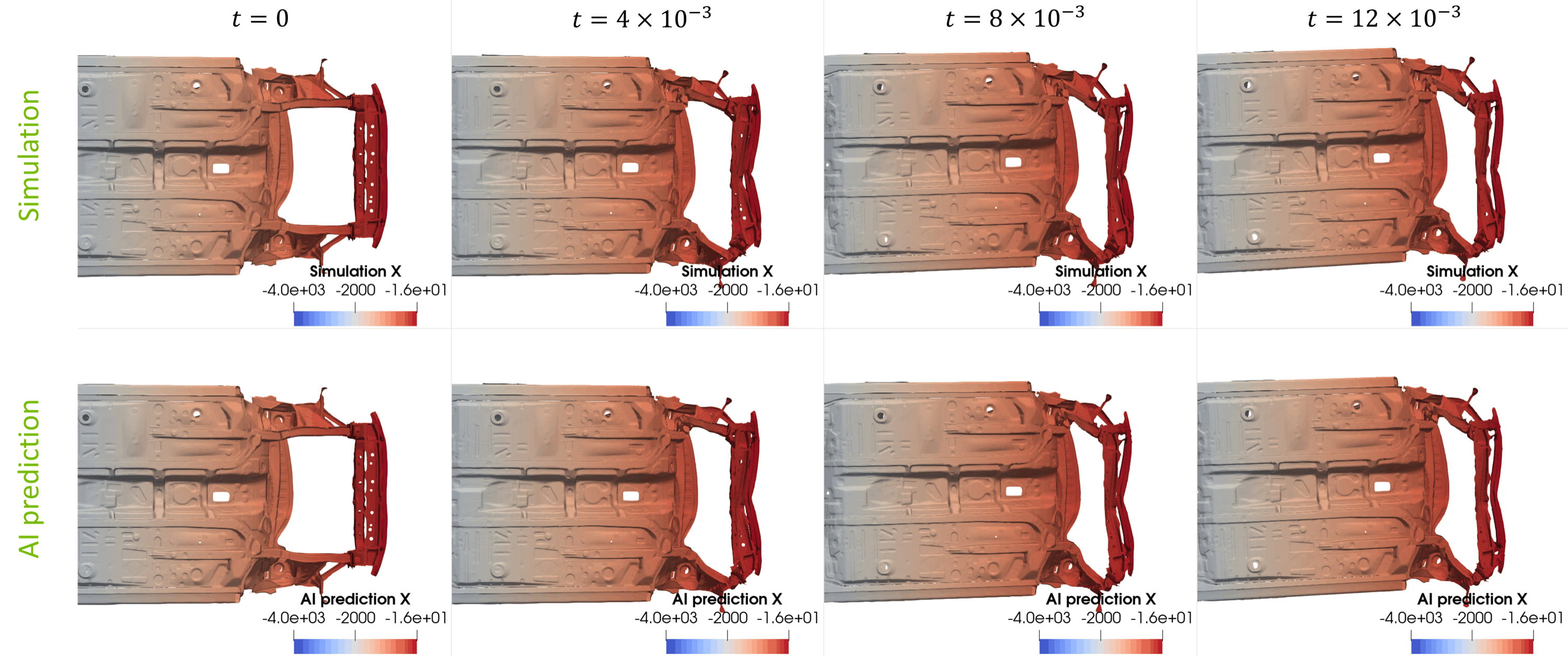}
        \caption{Car crash on test geometry deformation with time.}
        \label{fig:car_wtime_b}
    \end{subfigure}
    \caption{Visual comparison of car crash deformation with time. In both (a) and (b), the top row shows the ground truth FEA simulation, while the bottom row displays the GeoTransolver prediction.\\
    Video comparisons are available at: \url{https://github.com/NVIDIA/physicsnemo/blob/main/examples/structural_mechanics/crash/README.md}}
    \label{fig:car_wtime}
\end{figure}

Figures~\ref{fig:car_final} and \ref{fig:car_wtime} provide visual comparisons of the deformation states for representative test samples in the car crash dataset. In both figures, the top rows illustrate the ground truth results obtained from industrial-grade FEA simulations, while the bottom rows display the corresponding predictions generated by GeoTransolver. Specifically, Fig.~\ref{fig:car_final} focuses on the final deformation states, while Fig.~\ref{fig:car_wtime} tracks the structural evolution over time. In both cases, GeoTransolver accurately captures the overall collapse patterns and energy-absorbing structural deformations, including the complex buckling of the engine compartment and front-end assembly as the impact progresses.

\begin{figure}[!ht]
    \centering
    \includegraphics[width=\linewidth]{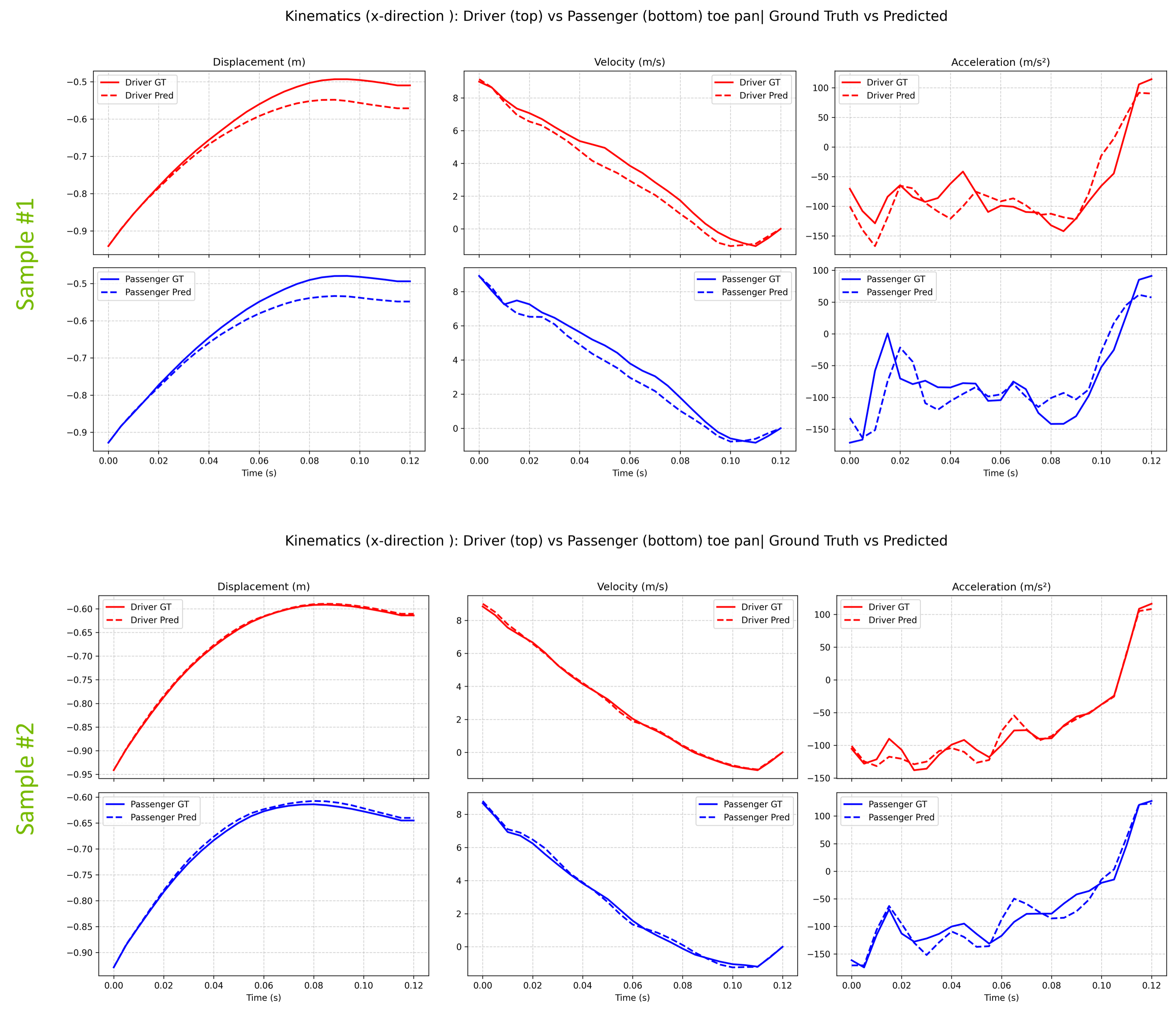}
    \caption{Temporal evolution of kinematics (x-direction) at the driver and passenger toe pan locations for car crash samples. Comparison between FEA ground truth and GeoTransolver prediction for position, velocity, and acceleration.}
    \label{fig:probe_kinematics}
\end{figure}

To further assess the fidelity of the temporal evolution, Fig.~\ref{fig:probe_kinematics} compares the kinematic profiles (position, velocity, and acceleration) predicted by GeoTransolver against FEA ground truth at the driver and passenger toe pan locations for test samples. The model demonstrates commendable accuracy in tracking the rapid deceleration profiles, effectively capturing the nonlinear transients that are critical for occupant safety assessment.

\begin{figure}[!ht]
    \centering
    \includegraphics[width=\linewidth]{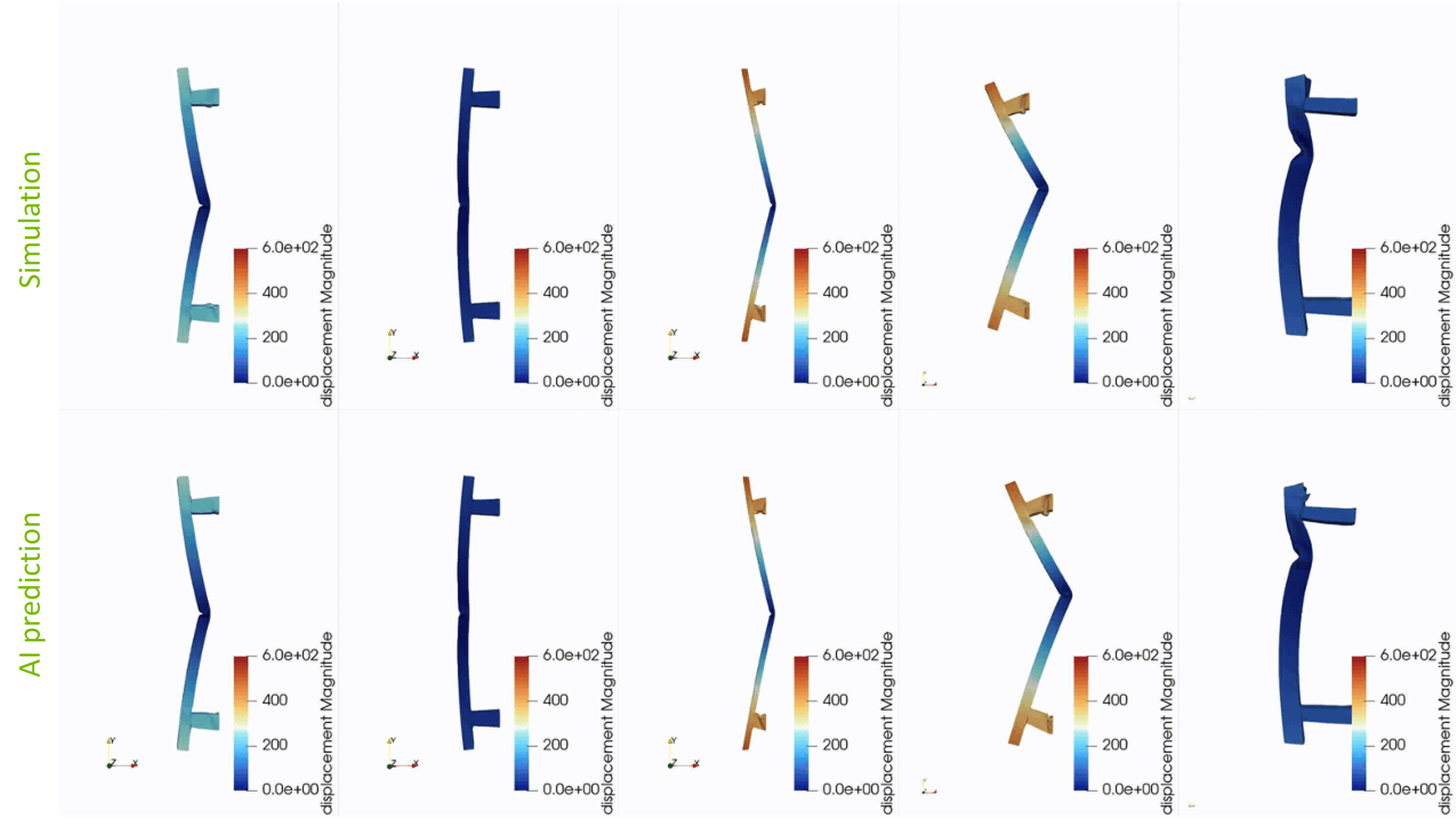}
    \caption{Visual comparison of final deformation states for five test samples in the bumper beam dataset. The top row shows the ground truth FEA simulation, while the bottom row displays the GeoTransolver prediction.}
    \label{fig:Bumper_final}
\end{figure}

Lastly, Fig.~\ref{fig:Bumper_final} illustrates the predictive performance on the bumper beam dataset across five distinct test samples. Similar to the Full-Vehicle results, the top row represents the ground truth simulations and the bottom row shows the AI-driven predictions. GeoTransolver consistently resolves the intricate folding and deformation modes of the bumper assembly under varying impact conditions, further validating the robustness and generalizability of the framework for industrial crash modeling.

\section{Conclusion}
Predicting the complex, highly nonlinear dynamics of automotive crash events is a critical challenge for surrogate modeling. In this work, we demonstrated that GeoTransolver \cite{adams2025geotransolver}, a geometry-aware operator learning framework, provides a viable, industrial-scale surrogate for high-fidelity crash dynamics prediction. Benchmarked on complex bumper beam and Full-Vehicle datasets, GeoTransolver accurately resolves plastic deformation patterns as well as position, velocity, and acceleration profiles at critical occupant locations, supporting its use as a practical accelerator for crashworthiness design workflows.

Our comparative analysis of temporal prediction strategies reveals that while autoregressive rollout suffers from cumulative instability over extended horizons, One-shot mapping emerges as a robust, computationally efficient alternative. Furthermore, time-conditional queries achieve the highest precision in resolving localized structural deformations. As a secondary contribution, we introduced a FLARE-based modification to the GeoTransolver attention backbone that reduces memory overhead by approximately $2\times$ while further improving predictive accuracy, yielding substantial error reductions in safety-critical metrics like occupant-area acceleration. Together, these findings underscore that geometry-aware operator learning—coupled with memory-efficient low-rank attention—offers a practical framework for accelerating high-fidelity design optimization in complex, safety-critical physical systems.

Building upon these results, future research will focus on extending GeoTransolver’s training across a more diverse range of automotive architectures and impact scenarios. By incorporating a broader spectrum of material properties and structural topologies, we aim to pave the way for a foundation model for nonlinear structural dynamics. Such a framework would provide a generalized predictive engine capable of zero-shot or few-shot transfer to novel vehicle designs, significantly reducing the reliance on computationally intensive FEA simulations across the industry.

\section*{Acknowledgements}
The author (D. Akhare) expresses his sincere gratitude to the \textbf{PhysicsNeMo} team at NVIDIA for their support and guidance, with special thanks to Ram Cherukuri, Rishikesh Ranade, and Srinivas Tadepalli. This research was conducted during an internship at NVIDIA.

\bibliographystyle{elsarticle-num}
\bibliography{ref}


\end{document}